\documentclass[letterpaper, 10 pt, conference]{ieeeconf}  

\IEEEoverridecommandlockouts                              

\usepackage{amsmath,amsfonts}
\usepackage{algorithmic}
\usepackage{algorithm}
\usepackage{array}
\usepackage{subcaption}
\usepackage{textcomp}
\usepackage{stfloats}
\usepackage{url}
\usepackage{verbatim}
\usepackage{tabularx}
\usepackage{graphicx}
\usepackage{cite}
\usepackage{booktabs} 
\usepackage{multirow} 
\usepackage{amsmath}
\usepackage{svg}
\usepackage{float}

\title{\LARGE \bf
Collective Learning for Unified Robotic Manipulation via Predictive Dynamics Adaptation
}

\author{
Shaocheng Liu$^{1}$, Yuan Meng$^{1}$, Zhenshan Bing$^{2,\dagger}$,
Alois Knoll$^{1}$%
\thanks{$^{1}$School of Computation, Information and Technology,
Technical University of Munich, Garching, Germany.
Email: shaocheng.liu@tum.de}%
\thanks{$^{2}$State Key Laboratory for Novel Software Technology and
the School of Science and Technology, Nanjing University
(Suzhou Campus), China.}%
\thanks{$^{\dagger}$Corresponding author: Zhenshan Bing,
\texttt{bing@nju.edu.cn}.}%
}
\begin{document}

\maketitle
\thispagestyle{empty}
\pagestyle{empty}

\begin{abstract}

Learning a collective policy for state-based Cartesian manipulation across diverse robotic embodiments presents a fundamental challenge due to embodiment-dependent transition dynamics. 
Collective learning aggregates multi-robot experience to discover shared task structure, yet naive aggregation can average incompatible action modes when different embodiments realize the same Cartesian command through different transition behavior.  This leads to strategy ambiguity and performance collapse on dynamically outlying embodiments. 
To address this, we propose PACE, a two-stage collective learning framework that separates shared task-space intent from embodiment-dependent transition dynamics. 
The first stage employs a representation layer pre-trained on offline trajectories to extract task-level context from collective data. The second stage introduces a decision layer equipped with a predictive dynamics adapter. 
By forcing the latent state to predict future transitions, this adapter acts as a latent regularizer that grounds the policy in the transition behavior of the executing robot while preserving real-time inference without online planning. 
Evaluations across simulated embodiments demonstrate that our method prevents performance collapse on distinct robots. It achieves over 93\% overall success and 89\% on dynamically out-of-distribution
zero-shot simulated robots, outperforming the Transformer behavior cloning (82\%) and diffusion policy baselines (66\%). Physical KUKA deployment further validates the feasibility of the full state-based pipeline under zero-shot sim-to-real transfer.
\end{abstract}

\section{Introduction}
\label{sec:intro}

Developing generalist robotic agents capable of mastering diverse manipulation tasks across heterogeneous embodiments remains a central objective in embodied AI. Recent large-scale robot learning efforts show that policies trained on collective data from many robots can acquire reusable task structure and improve transfer across platforms\cite{embodimentcollaboration2025openxembodimentroboticlearning,octomodelteam2024octoopensourcegeneralistrobot}.
Such collective data enables a policy to acquire reusable task structure beyond the experience of any individual robot. However, cross-embodiment collective learning also introduces heterogeneous sensing modalities, kinematic structures, and physical execution behaviors\cite{doshi2024scalingcrossembodiedlearningpolicy, wang2024scalingproprioceptivevisuallearningheterogeneous}
Achieving robust control across such heterogeneous embodiments therefore requires learning what can be shared across robots while preserving embodiment-specific factors that affect execution.

Recent works increasingly train shared policies on collective robot data to overcome the limitations of isolated single-robot training and to discover reusable structures across embodiments\cite{muller2024openbotfleet, wang2024robotfleetlearning}. Although this aggregation scales manipulation capabilities, dynamic heterogeneity remains a primary barrier to successful cross-embodiment transfer. Standard behavior cloning applied to these datasets typically treats state and action pairs as independent and identically distributed samples, inadvertently ignoring the physical context of the robot. For example, under a shared Cartesian command interface, a high-inertia arm and a lightweight collaborative arm may realize the same commanded end-effector displacement through different transitions. Training a single policy on such diverse dynamics leads to strategy ambiguity. Because the optimization process minimizes the expected error by averaging conflicting actuation modes, it ultimately generates suboptimal actions \cite{wang2024scalingproprioceptivevisuallearningheterogeneous}. Fully realizing the potential of collective learning thus requires bridging these low-level dynamic gaps by isolating shared task semantics from specific hardware physics.
\begin{figure}
    \centering
    \includegraphics[width=1.0\linewidth]{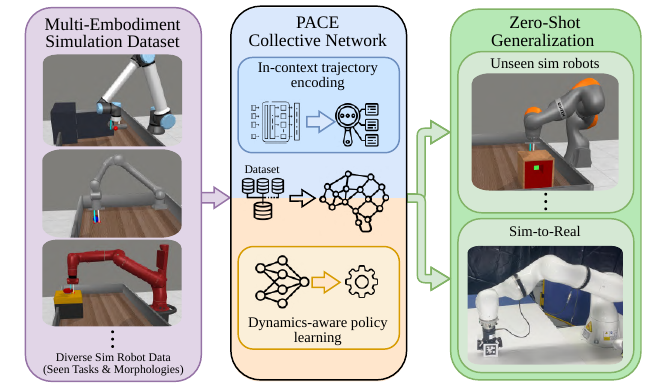}
    \vskip -.05in
    \caption{Overview of the PACE framework. The left panel shows the diverse multi-embodiment dataset. The central panel details the collective network, which consists of in-context trajectory encoding and dynamics-aware policy learning. The right panel demonstrates the cross-embodiment generalization to unseen simulated and real-world robots.}
    \vskip -.2in
    \label{fig:overview}
\end{figure}

Dynamics-aware collective learning benefits from separating what should be shared across robots from what should remain embodiment-aware. To achieve this, we propose a two-stage collective learning framework illustrated in Figure \ref{fig:overview}. The learning pipeline begins with a diverse multi-embodiment dataset covering various tasks and robot morphologies. The core collective network processes this data through two functional stages. First, a representation layer pre-trained on offline data extracts an embodiment-agnostic task context via masked trajectory reconstruction and contrastive learning \cite{wu2023maskedtrajectorymodelsprediction, lee2025classcontrastivelearningaction}. Second, a decision layer employs a predictive dynamics adapter as a latent regularizer to enable precise low-level control \cite{xie2025latentdiffusionplanningimitation}. By forcing the latent state to predict future transitions, the policy implicitly adapts to the physical constraints of the current robot. This design ultimately yields robust cross-embodiment generalization, allowing the policy to control unseen simulated robots and physical hardware. Furthermore, it avoids the computational latency associated with the online simulation of traditional world models \cite{hansen2024tdmpc2scalablerobustworld, hafner2024masteringdiversedomainsworld}. As a result, the system executes in real time while achieving an overall success rate of over 93\% and maintaining an 89\% success rate on challenging robots with out-of-distribution dynamics, effectively preventing the severe performance collapse observed in the Transformer behavior cloning (73\%) and diffusion (66\%) baselines.

The main contributions of our study are three folds:
\begin{itemize}
    \item We formulate dynamics-aware collective manipulation under a shared state-based Cartesian interface, where multi-robot data provides shared task structure while embodiment-dependent transition behavior creates strategy ambiguity.
    \item  We introduce a dual-objective latent regularizer that utilizes reward prediction to anchor the representation in task semantics and applies auxiliary next-step prediction to enforce adaptation across diverse manipulators.
    \item We conduct extensive empirical evaluations demonstrating that the proposed method prevents performance collapse on outlier robots, outperforms standard behavior cloning baselines, and enables zero-shot transfer capabilities on previously unseen physical hardware.
\end{itemize}
We name our framework as \textbf{P}redictive \textbf{A}daptation for \textbf{C}ollective \textbf{E}mbodiments Learning (PACE).


\section{Related Work}

Recent studies rely on collective and fleet learning to train models across diverse robotic platforms \cite{muller2024openbotfleet, wang2024robotfleetlearning, liu2024multitaskinteractive}. Early large-scale parallel data collection systems, which used arrays of identical manipulators to autonomously learn shared grasping skills, established the foundation for this approach \cite{levine2016learning}. Modern frameworks expand on this concept by aggregating experiences across multiple hardware instances, allowing robots to learn generalizable manipulation skills without isolated training. Recent cross-embodiment models, for example, train generalist agents by processing extensive visual observations combined with natural language instructions \cite{embodimentcollaboration2025openxembodimentroboticlearning, octomodelteam2024octoopensourcegeneralistrobot}. In parallel, hardware-centric approaches use universal interfaces to standardize the physical control space \cite{chi2024universalmanipulationinterfaceinthewild}. To handle structural differences within multi-robot datasets, current architectures align disparate state spaces into a shared latent space \cite{wang2024crossembodimentrobotmanipulationskill, chen2024miragecrossembodimentzeroshotpolicy} or apply heterogeneous pre-trained transformers \cite{wang2024scalingproprioceptivevisuallearningheterogeneous, bohlinger2025policyrunallendtoend}. Although these methods provide a strong basis for semantic generalization across robot fleets, they frequently struggle to solve the low-level dynamic discrepancies that arise when transferring policies to hardware with significantly different inertial properties.

To model action distributions, recent studies adopt sequence modeling and predictive architectures. Masked trajectory models reconstruct sequences of states and actions to learn representations applicable to control \cite{wu2023maskedtrajectorymodelsprediction}. Advanced autoregressive and diffusion frameworks further refine the generation of trajectories \cite{zhang2026chainofactiontrajectoryautoregressivemodeling, davies2025tenmarobustcrossembodimentrobot}, while contrastive learning disentangles task intent from morphological noise \cite{lee2025classcontrastivelearningaction}. Concurrently, model-based reinforcement learning captures transitions of the environment through explicit predictive world models \cite{hansen2024tdmpc2scalablerobustworld, hafner2024masteringdiversedomainsworld, tian2024predictiveinversedynamicsmodels}. Although explicit world models can capture system dynamics, they typically rely on iterative internal simulation for online planning. This mechanism introduces substantial computational latency during execution and requires active environmental interaction to adapt to new physical properties. Consequently, the application of explicit world models is fundamentally restricted in zero-shot cross-embodiment transfer, where policies must execute in real time from strict offline demonstrations. The proposed framework avoids this limitation by utilizing predictive dynamics exclusively as a latent regularizer, thereby enabling real-time physical adaptation without the computational overhead of online planning.

\section{Methodology}
\begin{figure*}[ht]
  \centering
  \includegraphics[width=1.0\textwidth]{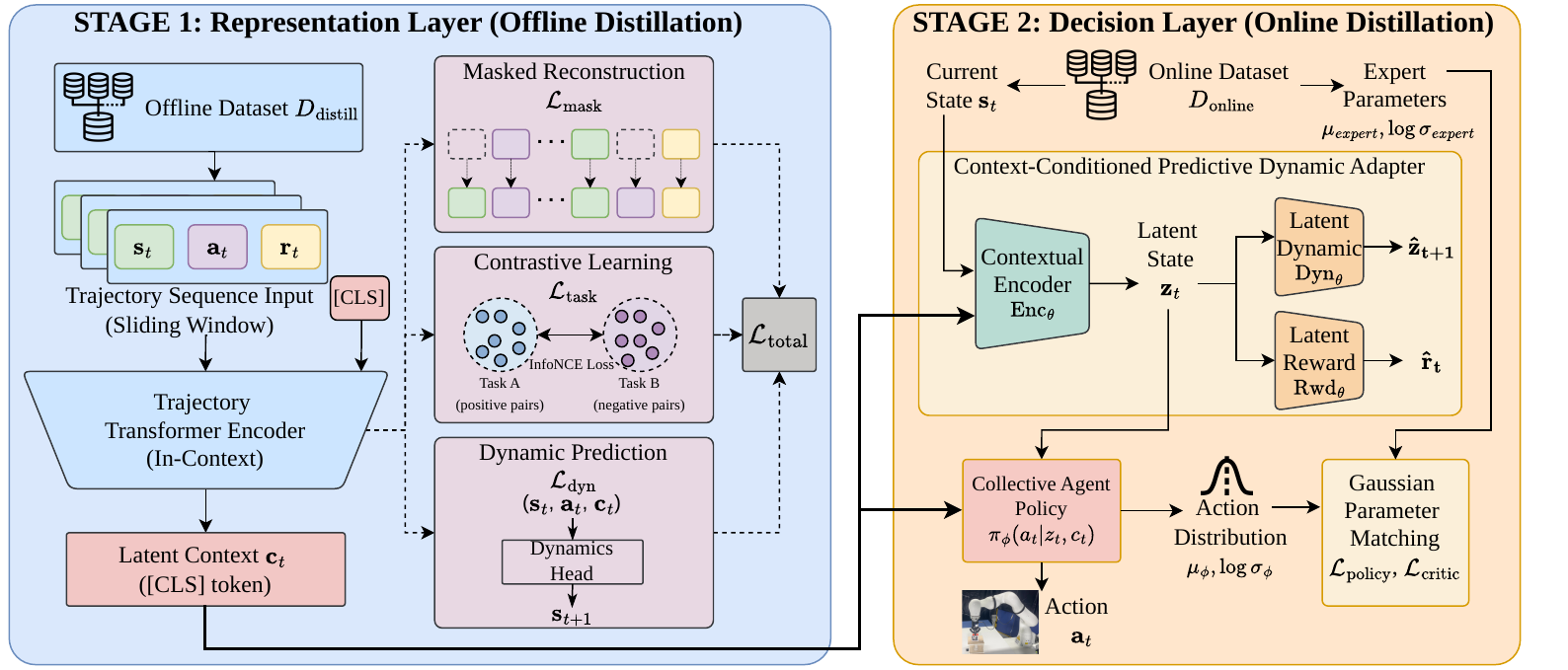}
\vskip -.1in
  \caption{\textbf{Structure of PACE.} The two-stage distillation framework features 1. an offline representation layer with a trajectory transformer encoder, and 2. an online decision layer with a predictive dynamics adapter and collective agent policy.}
\vskip -.3in
  \label{fig:method}
\end{figure*}
\subsection{Problem formulation}
\label{sec:problem_formulation}

We formulate collective multi-task imitation learning as a family of Markov Decision Processes (MDPs), $\mathcal{M} = \{M_{rob, task} \mid rob \in \mathcal{R}, \tau \in \mathcal{T}\}$, where $\mathcal{R}$ represents distinct robot embodiments and $\mathcal{T}$ represents manipulation tasks. 

\vspace{1mm}
\noindent \textbf{Unified state and action spaces.}
We use a unified Cartesian space to facilitate transfer. The state $\mathbf{s}_t \in \mathcal{S}$ describes the workspace geometry: $\mathbf{s}_t = [\mathbf{x}_{\text{tool}}, \mathbf{x}_{\text{obj}}, \mathbf{g}_{\text{pos}}] \in \mathbb{R}^{d_s}$, where $\mathbf{x}_{\text{tool}} \in \mathbb{R}^4$ comprises the end-effector position and gripper status, $\mathbf{x}_{\text{obj}}$ is the object state, and $\mathbf{g}_{\text{pos}}$ is the goal.This formulation helps to separate geometric reasoning from robot-specific kinematics.
Similarly, the action space $\mathcal{A}$ is standardized. We control end-effector via $\mathbf{a}_t = [\Delta x, \Delta y, \Delta z, a_{\text{grip}}]$, representing the 3D displacement and gripper actuation.

\vspace{1mm}
\noindent \textbf{Heterogeneous dynamics and objective.} 
Although $\mathcal{S}$ and $\mathcal{A}$ are unified, the transition dynamics $P_r(\mathbf{s}_{t+1} | \mathbf{s}_t, \mathbf{a}_t)$ depend explicitly on the robot. Due to varying physical properties (e.g., inertia, friction), the same action yields different transitions across robots (e.g., $P_{\text{Sawyer}} \neq P_{\text{Gen3}}$). 
Our objective is to learn a single collective policy $\pi_\theta(\mathbf{a}_t | \mathbf{s}_t)$ that implicitly adapts to these dynamics, maximizing the expected cumulative reward $\mathbb{E}[\sum \gamma^t R_{task}]$ across all pairs in $\mathcal{R} \times \mathcal{T}$.

\vspace{1mm}
\noindent \textbf{Origins of dynamic heterogeneity.}
Origins of dynamic heterogeneity. In this work, dynamic heterogeneity denotes embodiment-dependent transition behavior under the shared Cartesian command interface. This behavior arises from the interaction between robot physical properties, low-level tracking dynamics, and the controller abstraction used in simulation. Meta-World employs a motion capture (Mocap) controller commanding the end-effector to track a Cartesian pose via soft constraints, abstracting away joint torque generation. The simulator still enforces rigid-body dynamics governed by $M(q)\ddot{q}+C(q,\dot{q})\dot{q}+G(q)=\tau$. As robot inertial properties and tracking responses vary, identical Cartesian commands can yield different realized transitions. This controller-mediated transition shift renders the unified action space inconsistent without adaptive correction.

\subsection{System overview}
This two-stage framework (Fig. \ref{fig:method}) distills heterogeneous expert policies into a single dynamics-aware collective agent. Stage 1 extracts task-level context from collective trajectories, capturing the shared structure that should transfer across embodiments. Stage 2 uses predictive latent dynamics to preserve the embodiment-dependent transition behavior needed for reliable control. This design allows the policy to exploit multi-robot data while avoiding the action averaging that causes strategy ambiguity in pooled behavior cloning.

\subsection{Representation Layer: In-context trajectory encoding}
\label{sec:representation_layer}

The primary challenge is ``Strategy Ambiguity'': raw geometric states reveal neither the underlying robot dynamics nor the specific expert strategy. This layer therefore encodes heterogeneous trajectory history into a task-centered latent context that can be shared across embodiments.

\noindent \textbf{Architecture and inference procedure.}
We employ a Transformer-based architecture operating on a sliding window of the unified trajectory. Given an input sequence $\mathbf{h}_t = \{(\mathbf{s}_{t-H+1}, \mathbf{a}_{t-H+1}, \mathbf{r}_{t-H+1}), \dots, (\mathbf{s}_t, \mathbf{a}_t, \mathbf{r}_t) \}$, we prepend a learnable [CLS] token to aggregate global context. The Transformer processes this sequence via self-attention, outputting the [CLS] embedding $\mathbf{c}_t$. This vector acts as a static context, informing downstream modules of the robot identity and task nature.

To ensure $\mathbf{c}_t$ captures these properties, we train the encoder using a multi-objective loss on the dataset $\mathcal{D}_{\text{distill}}$:

\vspace{1mm}
\noindent \textbf{Temporal understanding via masked reconstruction ($\mathcal{L}_{\text{rec}}$).}
Following BERT, we enforce learning of temporal structure by reconstructing masked trajectory segments. We randomly mask states and actions with probability $p_{\text{mask}}$; the model must predict original values from the unmasked context. This self-supervised objective captures causal dependencies within the motion:
\begin{equation}
\mathcal{L}_{\text{rec}} = \frac{1}{|M|} \sum_{j \in M} | y_j - \hat{y}_j |^2
\end{equation}
where $M$ is the set of masked indices, $y_j$ is the ground truth, and $\hat{y}_j$ is the reconstruction.

\vspace{1mm}
\noindent \textbf{Task disentanglement via contrastive learning ($\mathcal{L}_{\text{task}}$).}
To distinguish semantic objectives within the multi-task dataset, we apply a supervised InfoNCE loss to $\mathbf{c}$. Trajectories from the same task are treated as positive pairs, while those from different tasks are negatives. This clusters the latent space primarily by task intent:
\begin{equation}
\mathcal{L}_{\text{task}} = - \log \frac{\exp(\text{sim}(\mathbf{c}_i, \mathbf{c}_j)/\tau)}{\sum_{k=1}^{N} \exp(\text{sim}(\mathbf{c}_i, \mathbf{c}_k)/\tau)}
\end{equation}
where $\mathbf{c}_i, \mathbf{c}_j$ are embeddings from the same task, and $\tau$ is the temperature.

\vspace{1mm}
\noindent \textbf{Physical grounding via auxiliary dynamics prediction ($\mathcal{L}_{\text{dyn}}$).}
To capture robot-specific transition dynamics, we attach a lightweight dynamics head predicting $\mathbf{s}_{t+1}$ given $\mathbf{s}_t, \mathbf{a}_t$, and $\mathbf{c}$. Since inputs are structurally unified, the burden of distinguishing robots falls entirely on $\mathbf{c}$. Minimizing this error encourages the [CLS] token to capture dynamics-related features necessary for transition prediction:
\begin{equation}
\mathcal{L}_{\text{dyn}} = \frac{1}{T-1} \sum_{t=0}^{T-2} | \mathbf{s}_{t+1} - f_{\text{dyn}}(\mathbf{s}_t, \mathbf{a}_t, \mathbf{c}) |^2_2
\end{equation}
\vspace{1mm}
\noindent \textbf{Total objective.}
The final loss is a weighted sum: $\mathcal{L}_{\text{total}} = \lambda_1 \mathcal{L}_{\text{mask}} + \lambda_2 \mathcal{L}_{\text{task}} + \lambda_3 \mathcal{L}_{\text{dyn}}$. This yields a representation encoding both task and environmentrequirements, serving as input for the Decision Layer.

\begin{figure*}[hb!]
    \vskip -.15in
    \centering
    \includegraphics[width=1\linewidth]{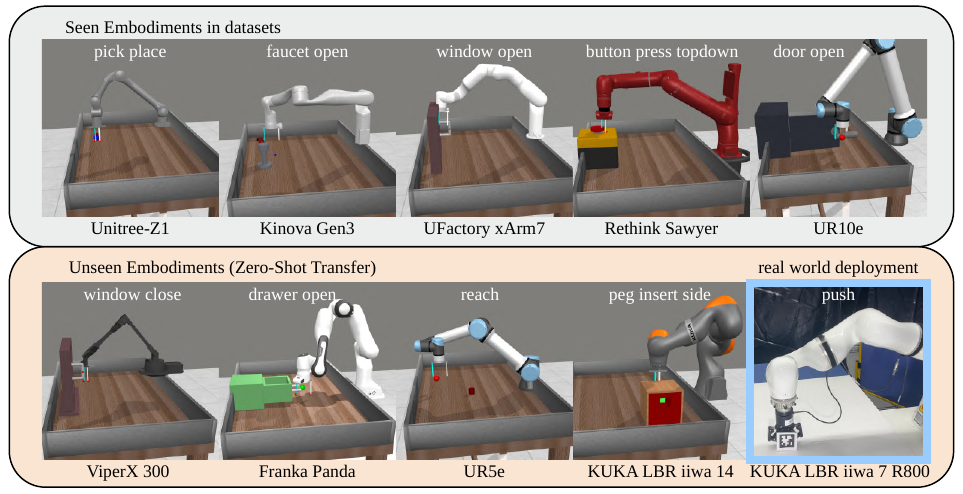}
    \vskip -.1in
    \caption{Cross-embodiment zero-shot transfer across diverse manipulation tasks. Physical specifications are in Table \ref{tab:robot_specs}. For more details, refer to our supplementary video.}
    \label{fig:robot photos}
\end{figure*}
\subsection{Decision Layer: Dynamics-aware policy learning}
\label{sec:decision_layer}

While the Representation Layer extracts a semantic context $\mathbf{c}_t$, it lacks the explicit physical grounding required to control diverse embodiments (e.g., differentiating high-inertia industrial arms from lightweight cobots). The Decision Layer bridges this gap using a Predictive Dynamics Adapter to refine the latent state $\mathbf{z}_t$.

Unlike standard World Models \cite{hafner2024masteringdiversedomainsworld} \cite{hansen2024tdmpc2scalablerobustworld} that use learned dynamics for planning or synthetic rollouts, we use the predictive model strictly as a representation regularizer. By minimizing the next-step prediction error $\mathcal{L}_{dyn}$ in latent space, we force $\mathbf{z}_t$ $z_t$ to encode robot-specific transition dynamics within a latent world model framework. This allows the policy $\pi_\phi(a_t|z_t, c_t)$ to remain model-free during inference, avoiding online planning latency while retaining the benefits of physics-aware representations.

Our Predictive Dynamics Adapter operates alongside a reward prediction module. The reward model ensures the latent space retains task-critical semantic features. The dynamics model simultaneously adapts the representation to specific physical transition delays. This dual structure prevents the dynamics objective from overshadowing the core manipulation intent.

\vspace{1mm} \noindent \textbf{Dynamics-aware latent encoding}. This module optimizes a predictive transition model $(\text{Dyn}_\theta)$ and a reward model $(\text{Rwd}_\theta)$ to get a physics-grounded latent space:

\textit{1. Contextual encoder ($\text{Enc}_\theta$).}
The encoder maps the raw state $\mathbf{s}_t$ and frozen task embedding $\mathbf{c}_t$ to a latent state $\mathbf{z}_t$:
\vskip -.1in
\begin{equation}
\mathbf{z}_t = \text{Enc}_{\theta}(\mathbf{s}_t, \mathbf{c}_t)
\label{eq:encoder}
\end{equation}
\vskip -.05in

\textit{2. Latent dynamics model ($\text{Dyn}_{\theta}$).}
This predicts the next latent state $\hat{\mathbf{z}}_{t+1}$ from $\mathbf{z}_t$ and $\mathbf{a}_t$. Minimizing transition error encodes the robot-specific dynamics required for prediction into $\mathbf{z}_t$:
\vskip -.1in
\begin{equation}
\hat{\mathbf{z}}_{t+1} = \text{Dyn}_{\theta}(\mathbf{z}_t, \mathbf{a}_t, \mathbf{c}_t)
\label{eq:dynamics_model}
\end{equation}
\vskip -.05in
\textit{3. Reward model ($\text{Rew}_{\theta}$).}
We predict the immediate reward $\hat{r}_t$ to ensure $\mathbf{z}_t$ captures task-relevant features, mapping the expectation of a categorical distribution via a symlog function:
\vskip -.1in
\begin{equation}
\hat{r}_t = \text{symexp}(\mathbb{E}[\text{Rew}_{\theta}(\mathbf{z}_t, \mathbf{a}_t, \mathbf{c}_t)])
\label{eq:reward_model}
\end{equation}
\vskip -.05in

\vspace{1mm}
\noindent \textbf{Unified collective policy.} The collective policy $\pi_\phi(\mathbf{a}_t | \mathbf{z}_t, \mathbf{c}_t)$ acts as a unified controller. Since conditioning on $(\mathbf{z}_t, \mathbf{c}_t)$ resolves robot identity and task intent, the mapping to action becomes more consistent. We thus employ Gaussian Parameter Matching for distillation, minimizing the MSE between student and expert parameters:
\begin{equation}
\mathcal{L}_{\text{policy}} = \| \mu_{\phi} - \mu_{\text{expert}} \|^2 + \lambda \cdot \| \log \sigma_{\phi} - \log \sigma_{\text{expert}} \|^2
\end{equation}
where $\mu$ and $\log \sigma$ are the mean and log-variance of the action distribution. MSE regression provides better numerical stability than KL divergence for these deterministic expert policies, ensuring the agent replicates the embodiment-specific strategy.

\section{Experiments}
This section evaluates whether PACE improves collective manipulation learning across robot embodiments with different transition behavior. We use a shared low-dimensional state representation and Cartesian action interface for all methods, so the comparison focuses on embodiment-dependent differences in how commanded actions are realized.

Preliminary observations highlight a critical limitation in standard Behavior Cloning baselines. They achieve high success rates on lightweight manipulators but degrade on embodiments with substantial physical discrepancies. We identify two primary failure modes. First, baselines struggle to control seen outliers with distinct kinematics, failing to model heterogeneous dynamics. Second, this limitation restricts zero-shot transferability, preventing adaptation to unseen robots with heavy dynamics.

We therefore investigate whether auxiliary dynamics learning can effectively regularize cross-embodiment adaptation. Our analysis focuses on three core questions:
\begin{itemize}
    \item \textbf{RQ1 (Robustness to heterogeneity):} Can the Latent Dynamics Regularizer prevent performance degradation on outlier embodiments?
    \item \textbf{RQ2 (Mechanism of alignment):} How does the learned representation facilitate this adaptability?
    \item \textbf{RQ3 (Data efficiency):} Does modeling environmental transitions improve sample efficiency?
\end{itemize}

\subsection{Experimental setup}

\paragraph{Environments and embodiments} 
We evaluate the framework on our customized Meta-World~\cite{yu2019meta} benchmark. 
The evaluation includes 10 manipulation tasks inherited from Meta-World, illustrated in Fig.~\ref{fig:robot photos}. 
Specifically, we incorporate 9 robot meshes derived from the MuJoCo Menagerie~\cite{menagerie2022github}, whose dynamic and physical properties are summarized in Table~\ref{tab:robot_specs}.
When combined with different robot embodiments, this results in a total of 90 distinct task scenarios.
For sim-to-real transfer validation, we deploy a KUKA LBR iiwa 7 R800 robotic arm equipped with a Robotiq 2F-85 gripper. A RealSense D455 camera is mounted at the edge of the table to provide visual perception.
We use AprilTag markers to obtain precise 3D object positions.

\begin{table}[htbp]
\centering
\small
\setlength{\tabcolsep}{2pt}
\caption{Physical specifications of evaluated robotic manipulators.}
\label{tab:robot_specs}
\begin{tabular}{lccc}
\toprule
\textbf{Robot Model} & \textbf{Mass(kg)} & \textbf{Payload(kg)} & \textbf{Reach(mm)} \\
\midrule
\multicolumn{4}{c}{\textit{Seen Embodiments (Training \& Testing)}} \\
\midrule
Unitree Z1 & 4.5 & 3.0 & 740 \\
Kinova Gen3 & 8.2 & 4.0 & 902 \\
UFactory xArm7 & 13.7 & 3.5 & 700 \\
Rethink Sawyer & 19.0 & 4.0 & 1260 \\
UR10e & \textbf{33.5} & 12.5 & 1300 \\
\midrule
\multicolumn{4}{c}{\textit{Unseen Embodiments (Zero-Shot Transfer)}} \\
\midrule
ViperX 300 & 3.6 & 0.75 & 750 \\
Franka Panda & 18.0 & 3.0 & 855 \\
UR5e & 20.6 & 5.0 & 850 \\
KUKA LBR iiwa 14 & \textbf{29.9} & 14.0 & 820 \\
\midrule
\multicolumn{4}{c}{\textit{Real World Depolyment (Zero-Shot Transfer)}} \\
\midrule
KUKA LBR iiwa 7 R800 & 23.9 & 7.0 & 800 \\
\bottomrule
\end{tabular}
\vskip -.1in

\end{table}

\paragraph{Data collection and training}
We generate a dataset using 50 expert policies. The training pipeline consists of two phases: (1) distilling data from expert training ($\mathcal{D}_{\text{distill}}$) to pre-train the representation layer, and (2) employing online distillation ($\mathcal{D}_{\text{dagger}}$) to collect on-policy data for fine-tuning.

\paragraph{Embodiment scaling protocol}
To investigate the scaling properties of generalization (\textbf{RQ3}), we construct training subsets with $N=\{1, 3, 5\}$ robots. We analyze whether performance on outliers improves with embodiment diversity or encounters a generalization ceiling.

\paragraph{Baselines \& metrics}
We evaluate our approach against two competitive baselines to guarantee a strict comparison. 
Section II provides detailed theoretical justifications for these exclusions. Our first baseline is (1) a joint-training transformer model. This baseline integrates a Transformer encoder and an action inference agent. It predicts actions from a sliding window of states. This structure matches the exact historical field of view of our proposed model. The agent module fits the expert action distribution directly. It lacks auxiliary context or dynamic regularization. Our second baseline is (2) a Diffusion Policy (DP). This model represents the current state-of-the-art for multimodal imitation learning. It trains on the exact same offline datasets and predicts continuous action chunks. It relies entirely on implicit task discrimination without explicit physical alignment. We measure performance using the Success rate across all task-robot pairs over 5 seeds, reporting the mean values and standard deviations for comparison ($\mu\pm\sigma$).

\begin{figure*}[tp]
    \centering
    \begin{subfigure}[b]{0.6\textwidth}
        \centering
        \includegraphics[width=\textwidth]{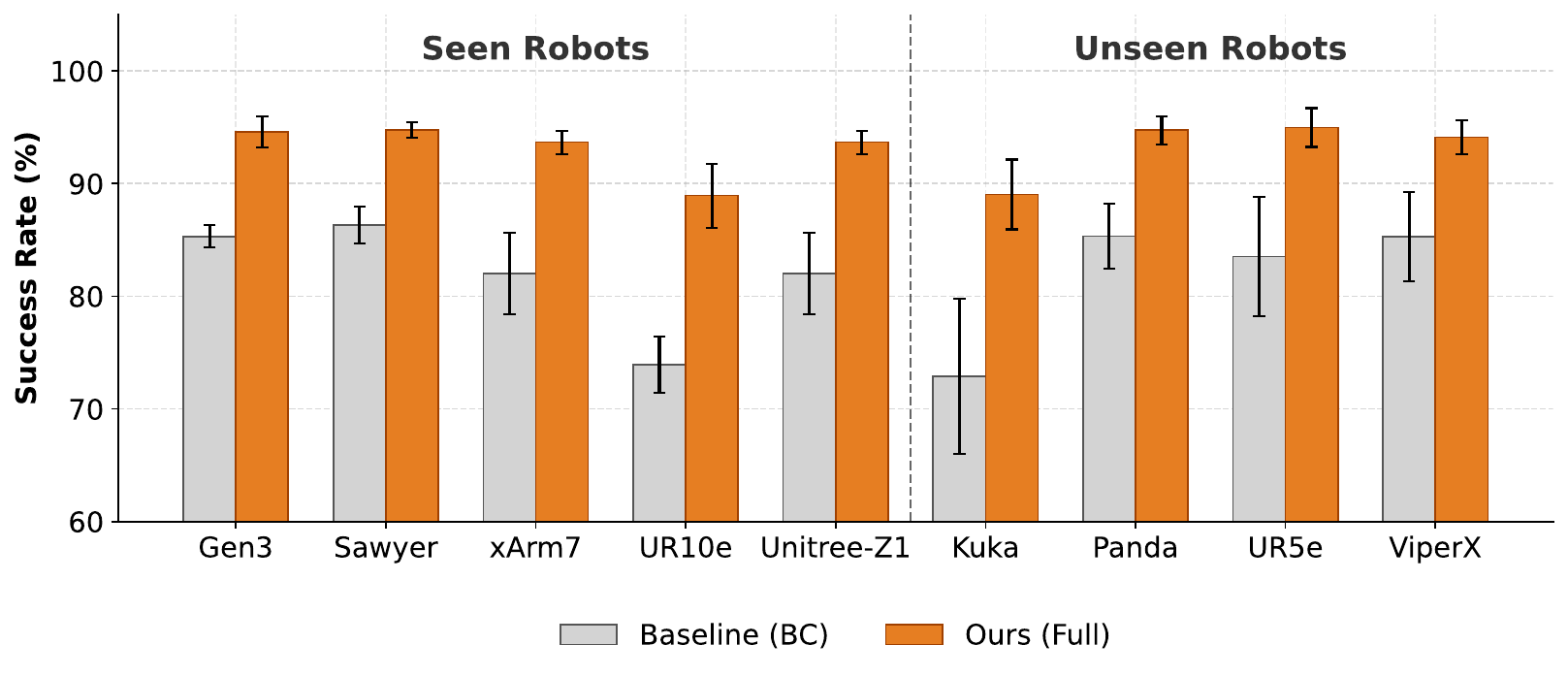}
        \caption{Comparison of per-robot success rates.}
        \label{fig:bar_comparison}
    \end{subfigure}
    \hfill
    \begin{subfigure}[b]{0.38\textwidth}
        \centering
        \includegraphics[width=\textwidth]{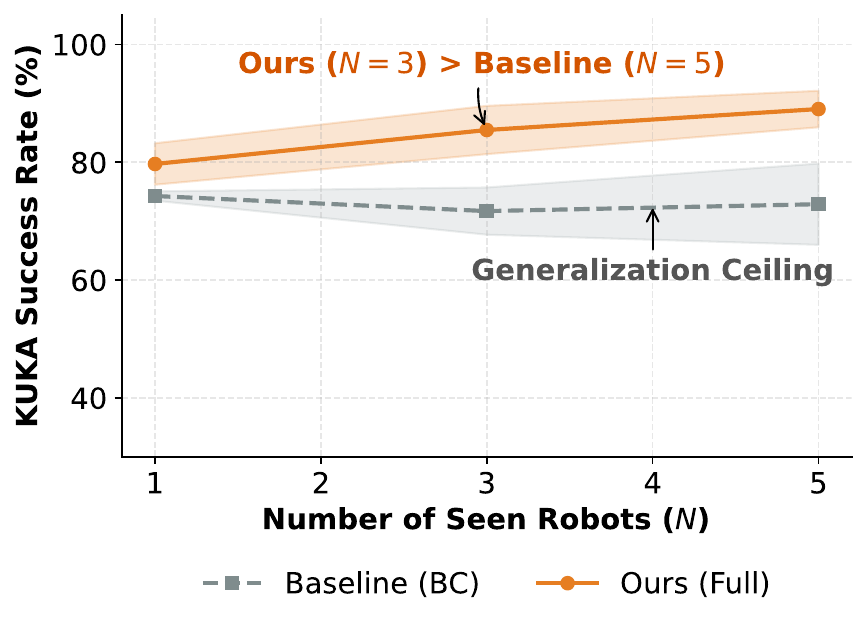}
        \caption{Generalization scaling.}
        \label{fig:scaling_law}
    \end{subfigure}
    \vskip -.05in
    \caption{\textbf{Performance Analysis.} (a) Per-robot success rate. Our method prevents performance collapse on outliers. (b) Success rate versus the number of seen robots ($N$).}
    \vskip -.2in
    \label{fig:main_performance}
\end{figure*}

\subsection{Robustness to dynamic heterogeneity}
\label{sec:results_main}

We first address \textbf{RQ1} by evaluating performance across seen and unseen embodiments. Table \ref{tab:main_results} summarizes the aggregated success rates, while Figure \ref{fig:main_performance} provides a detailed breakdown.

\begin{table}[h]
\centering
\vskip -.05in
\small 
\setlength{\tabcolsep}{4pt} 
\caption{\textbf{Generalization Performance.} Success rates (\%) on seen ($N=5$) and unseen ($N=4$) robots.}
\vskip -.05in
\label{tab:main_results}
\begin{tabularx}{\columnwidth}{Xcc} 
\toprule
\textbf{Method} & \textbf{Seen (\%)} & \textbf{Unseen (\%)} \\
\midrule
Transformer BC & 82.66 $\pm$ 5.16  & 81.75 $\pm$ 5.99 \\
\midrule
Diffusion Policy (DP) &  66.40 $\pm$ 3.85 & 65.25 $\pm$ 4.12\\
\midrule
\textbf{PACE (Ours)} & \textbf{93.17 $\pm$ 2.41} & \textbf{93.21 $\pm$ 2.81} \\
\bottomrule
\end{tabularx}
\vskip -.1in
\end{table}

\paragraph{Overall performance and the heterogeneity gap} 

Table \ref{tab:main_results} shows that the proposed framework achieves a consistent success rate of over 93\% across all robot categories, outperforming the Transformer and DP baselines. Task-level analysis indicates the DP struggles primarily with pushing, pick-placing, and side peg insertion, while performing comparably on the remaining tasks. This deficit likely stems from data scarcity. Diffusion architectures require extensive datasets to model complex multi-modal distributions, making them less effective under a constrained distillation setup. The Transformer baseline and the proposed model avoid this limitation, demonstrating higher sample efficiency.

Although the Transformer baseline outperforms the DP, aggregate metrics mask the absence of a traditional zero-shot transfer gap. Average success rates on seen and unseen embodiments remain nearly identical for both models. This parity suggests unseen kinematic structures are not the primary generalization barrier. For instance, controlling a lightweight unseen robot like the ViperX ($\sim85\%$ transformer BC baseline success) is fundamentally simpler than controlling a heavy seen robot like the UR10e ($\sim74\%$ transformer BC baseline success) due to different dynamic characteristics. Evaluating robustness via a binary seen-unseen split therefore masks the core challenge of dynamic divergence.

The actual challenge is managing out-of-distribution dynamics at the tails of the training distribution. Focusing on outlier embodiments isolates dynamic properties from kinematic structures. As Figure \ref{fig:bar_comparison} illustrates, the Transformer baseline suffers a performance collapse on high-inertia manipulators, dropping to $\sim74\%$ on the seen UR10e and $\sim73\%$ on the unseen KUKA iiwa. This degradation shows standard behavior cloning fails to resolve conflicting kinematic gradients, outputting invalid averaged actions. The proposed framework prevents this collapse. It maintains over 93\% success on standard lightweight arms and improves heavy arm performance by over 15\% points, reaching $\sim89\%$ on both the UR10e and the KUKA iiwa. By acting as a latent regularizer, the predictive dynamics adapter grounds the control strategy in the physical reality of the hardware, overcoming heterogeneous dynamic structures.

\paragraph{Breaking the generalization ceiling} 

\begin{figure}[h]
    \vskip -.1in
    \centering
    \includegraphics[width=0.9\linewidth]{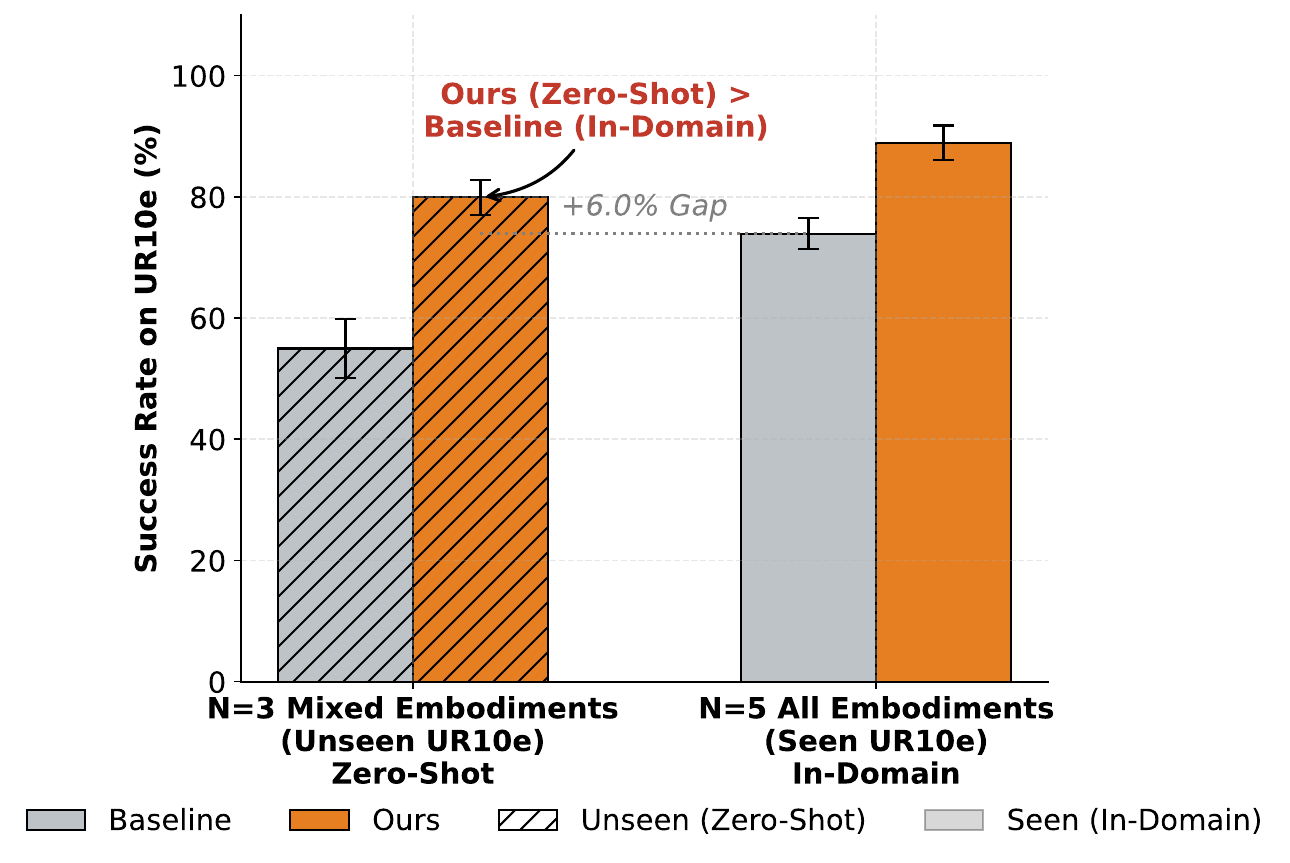}
    \vskip -.1in
    \caption{Zero-Shot vs. In-Domain generalization on UR10e across varying training set sizes.}
    \vskip -.2in
    \label{fig:ur10e}
\end{figure}

Figure \ref{fig:scaling_law} details generalization scaling on the held-out KUKA iiwa robot from a single-robot setting ($N=1$ Sawyer) to the full dataset ($N=5$). Introducing heterogeneity creates a critical divergence. The baseline achieves reasonable performance at $N=1$ but suffers negative transfer when expanded to $N=3$ (Sawyer, Unitree-Z1, Gen3), indicating an inability to solve conflicting kinematic gradients. The proposed method exhibits a positive scaling law, translating increased embodiment diversity into monotonic success rate gains at $N=3$ and $N=5$. Trained on just $N=3$ robots, the model surpasses the baseline trained on all five, confirming it mitigates interference and utilizes diverse physical priors. Analyzing the UR10e (Figure \ref{fig:ur10e}) isolates this out-of-distribution dynamics challenge. The lightweight-dominated $N=3$ training set creates a distribution shift against the heavy-duty UR10e. The baseline achieves only a 60\% success rate, overfitting the lightweight mode and failing to extrapolate. The proposed model attains 80\%, outperforming the baseline trained on the complete dataset containing the UR10e. This confirms the representation layer captures embodiment-specific temporal patterns rather than relying on mere data scaling.

\subsection{Analysis of Learned Representation}
\label{sec:results_rep}

\begin{figure}[t!]

    \centering
    \includegraphics[width=.9\linewidth]{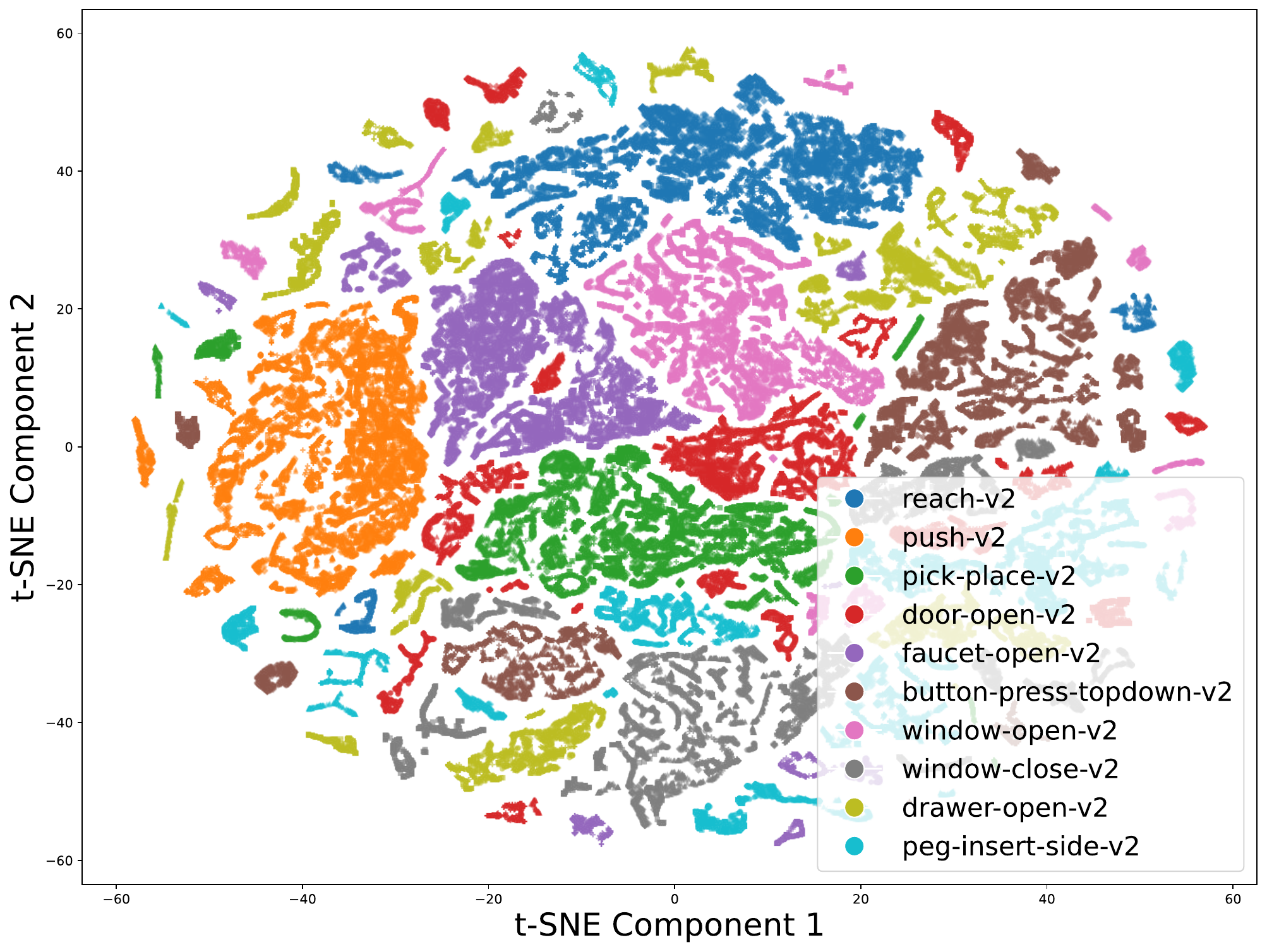}
    \caption{\textbf{t-SNE cluster Visualization.} t-SNE projection of context embeddings colored by task ID.}
    \vskip -.2in
    \label{fig:tsne}
\end{figure}

We analyze the geometric structure of the learned latent space to understand how it facilitates cross-embodiment transfer (\textbf{RQ2}). Fig.~\ref{fig:tsne} visualizes the t-SNE projection of the context embeddings $c_t$ across different tasks and robots.

\paragraph{Semantic clustering and functional equivalence} The visualization reveals distinct clusters organized primarily by task identifier rather than robot identifier. This supports the hypothesis of functional equivalence: the representation layer strongly emphasizes high-level semantic intent over low-level kinematic identifiers.

\paragraph{Latent dynamics encoding}
Visualizing outlier embodiments offers further insight. This suggests the encoder extracts semantic task intent while implicitly preserving essential dynamic signatures required for the second stage. The performance recovery on UR10e implies that $\mathbf{c}_t$ contains sufficient latent information for the dynamics adapter to infer dynamics-related characteristics.

\subsection{Ablation Studies}
\label{sec:results_ablation}

To address \textbf{RQ3}, we examine the contribution of specific components, focusing on the necessity of the Stage 2 dynamics adapter versus the Stage 1 representation objectives.

\paragraph{The Regularization effect of the latent dynamics adapter}
Our ablation results reveal a functional separation between the two stages, validating the necessity of each.

\begin{figure}[h]
\vskip -.1in
    \centering
    \includegraphics[width=1.0\linewidth]{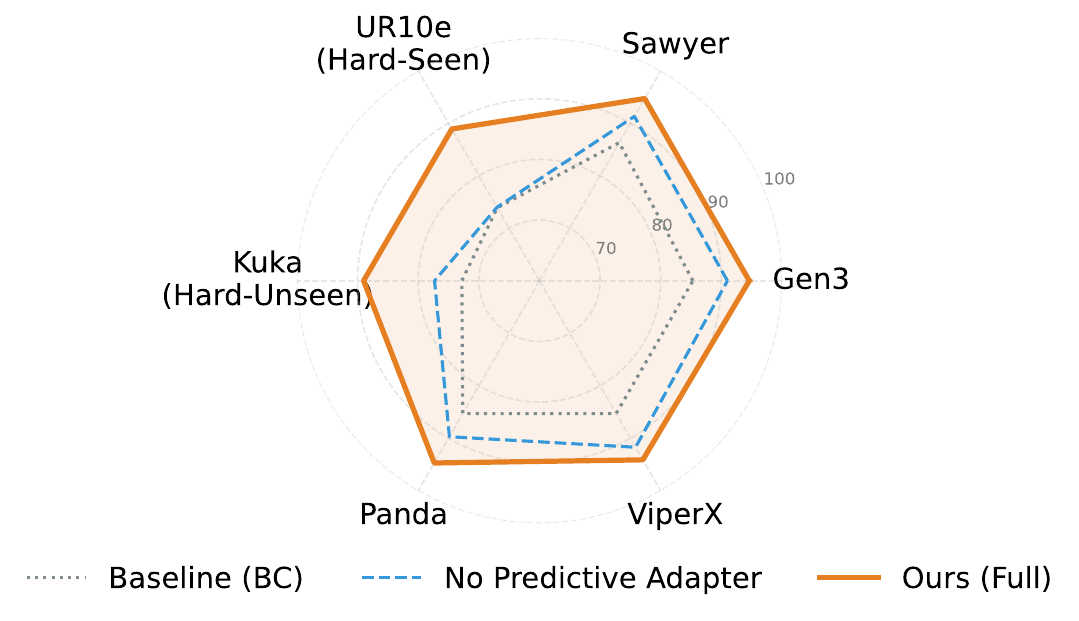}
    \vskip -.1in
    \caption{\textbf{Ablation Analysis.} Performance profile across 6 representative embodiments.}
    \label{fig:radar_ablation}
\end{figure}

\begin{itemize}
    \item \textbf{Stage 1 establishes intent:}
    As shown in Figure \ref{fig:radar_ablation}, the variant without the dynamics adapter (blue dashed line) outperforms the Transformer Baseline (gray dotted line) on standard lightweight manipulators, improving success rates from $\sim82\%$ to over $90\%$. The representation layer successfully extracts the task intent $c_t$. This extraction resolves the strategy ambiguity that causes the baseline models to fail. 
    
    \item \textbf{Stage 2 establishes dynamics:}
    However, Stage 1 alone is insufficient for physically distinct outliers. The policy without the dynamics adapter suffers a performance drop on high-inertia robots (e.g., UR10e success drops to $\sim74\%$). By introducing the predictive dynamics adapter (Stage 2), the Full Method (orange solid line) recovers this performance to $\sim89\%$.
\end{itemize}

Stage 1 identifies general task strategies to establish standard performance across all robots. Stage 2 then adapts to the specific dynamics of each hardware. We noticed in the laboratory that this second stage effectively corrects the performance drops on outlier equipment.

\paragraph{Impact of representation objectives} Table \ref{tab:ablation_stage1} shows removing the auxiliary dynamics loss in Stage 1 only slightly reduces performance. Stage 1 provides the necessary task context to prevent action conflicts. The latent dynamics adapter in Stage 2 then corrects the physical differences. The adapter successfully stabilizes the robot movements even when the initial static representations are imperfect.

\begin{table}[t!]
\caption{\textbf{Stage 1 Component Analysis.} Success rates (\%) with a frozen Stage 2 dynamics adapter.}
\vskip -.05in
\centering
\small
\setlength{\tabcolsep}{3pt}
\label{tab:ablation_stage1}
\begin{tabular}{l|ccc|cc}
\toprule
\multirow{2}{*}{\textbf{Configuration}} & \multicolumn{3}{c|}{\textbf{Objectives}} & \multicolumn{2}{c}{\textbf{Success Rate (\%)}} \\
 & $\mathcal{L}_{rec}$ & $\mathcal{L}_{task}$ & $\mathcal{L}_{dyn}$ & \textbf{Seen} & \textbf{Unseen} \\
\midrule
Full Method & \checkmark & \checkmark & \checkmark & \textbf{93.17} & \textbf{93.21} \\
\midrule
No $\mathcal{L}_{dyn}$ (S1) & \checkmark & \checkmark & - & 91.50 & 92.38 \\
No $\mathcal{L}_{NCE}$ (S1) & \checkmark & - & \checkmark & 90.64 & 91.38 \\
\bottomrule
\end{tabular}
\vskip -.2in
\end{table}

\begin{figure*}[t!]
    \centering
    \includegraphics[width=1\linewidth]{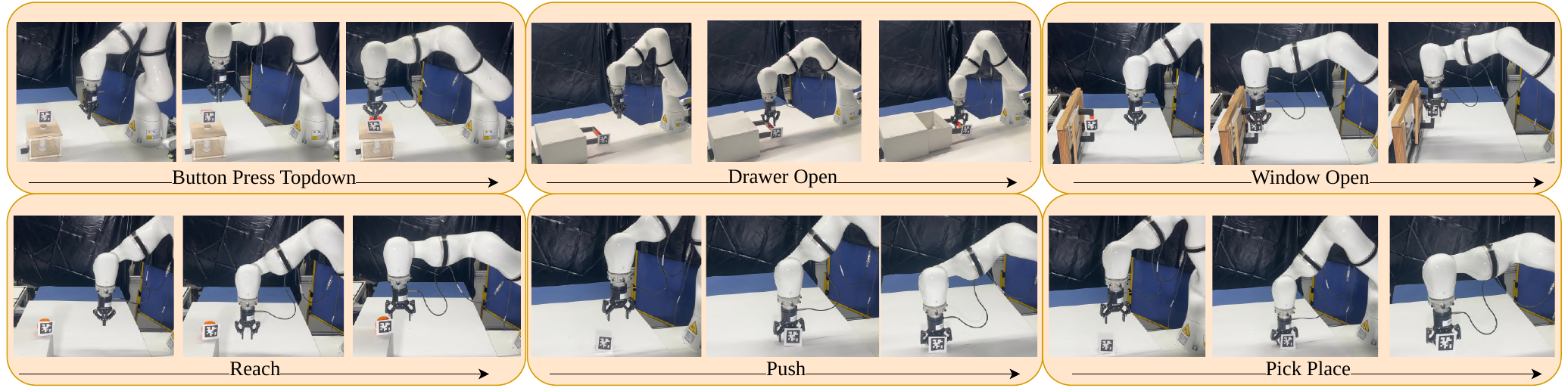}
    \vskip -.05in
    \caption{Qualitative demonstration of zero-shot sim-to-real transfer on the unseen KUKA iiwa robot. For more details, refer to our supplementary video.}
    \vskip -.2in
    \label{fig:deployment}
\end{figure*}

\subsection{Real world deployment}
To evaluate the practical viability and transferability of the proposed framework, we deploy the trained policy on a physical KUKA LBR iiwa 7 R800 platform. Real-world environments introduce substantial non-idealities absent in simulation, including unmodeled joint friction, inertial discrepancies, and inherent control latencies. These factors create a substantial domain gap that can destabilize standard cross-embodiment policies. In the real world, we conduct six manipulation tasks: Reach, Push, Pick-Place, Button Press, Window Open, and Drawer Open. 
\begin{table}[htbp]
\caption{Success rates and episode counts across manipulation tasks.}
\vskip -.05in
\centering
\small
\label{tab:real_results}
\begin{tabular}{lcc}
\toprule
\textbf{Task Name} & \textbf{Successes / 20 Eps} & \textbf{Success Rate (\%)} \\
\midrule
Reach         & 18 / 20 & 90.0 \\
Push          & 17 / 20 & 85.0 \\
Pick-Place    & 17 / 20 & 85.0 \\
Button Press  & 16 / 20 & 80.0\\
Drawer Open   & 14 / 20 & 70.0 \\
Window Open   & 13 / 20 & 65.0  \\
\midrule
\textbf{Average} & \textbf{15.8 / 20} & \textbf{79.2} \\
\bottomrule
\end{tabular}
\vskip -.1in
\end{table}

Table \ref{tab:real_results} summarizes the zero-shot physical deployment, showing an overall 79.2\% success rate. While the policy maintains high performance on free-space tasks, contact-rich tasks reveal distinct failure modes. A primary issue is the mechanical discrepancy between simulated and physical grippers. Direct action mapping causes unstable physical execution, corrupting state observations and inducing subsequent action errors. Heavily constrained tasks also expose specific vulnerabilities. In the Window Open task, the gripper frequently pushes against the handle side or frame—a trajectory that succeeds in simulation but fails physically. During the Drawer Open task, the real drawer tilts slightly downwards when pulled. The arm follows this unmodeled tilt, collides with the table, and triggers safety limits. Despite these hardware-specific issues, the agent completes most objectives by adapting to unmodeled execution latencies rather than rigidly replicating simulated trajectories. These results confirm that training across diverse simulated embodiments produces generalizable control policies. The predictive dynamics regularizer ensures robust zero-shot sim-to-real transfer without real-world fine-tuning.

\section{Conclusion}
This research introduces a two-stage collective learning framework to resolve dynamic heterogeneity in multi-robot manipulation. The system combines a representation layer for extracting task intent with a decision layer containing a predictive dynamics adapter, effectively decoupling semantic objectives from physical transition rules. Empirical evaluations show this predictive regularization prevents performance degradation on kinematic outliers and enables zero-shot transfer to physical hardware like the KUKA manipulator. Although a Cartesian action space prioritizes workspace geometry, this abstraction bypasses joint-level constraints and kinematic singularities, occasionally causing execution stalls. Future work should integrate joint state feedback directly into the predictive dynamics adapter to ensure joint-level compliance. Finally, broader quantitative testing in multi-tasks will help establish a rigorous understanding of system boundaries and deployment capabilities.

\section{Acknowledgment}
This research is supported by New Generation Artificial Intelligence-National Science and Technology Major Project (2025ZD0122903).

\bibliographystyle{IEEEtran}
\bibliography{reference}
\end{document}